\def\year{2022}\relax
\documentclass[letterpaper]{article} 
\usepackage{aaai22}  
\usepackage{times}  
\usepackage{helvet}  
\usepackage{courier}  
\usepackage[hyphens]{url}  
\usepackage{graphicx} 
\usepackage{natbib}  
\usepackage{caption} 
\DeclareCaptionStyle{ruled}{labelfont=normalfont,labelsep=colon,strut=off} 
\usepackage{dsfont}
\usepackage{subfig}
\usepackage{amsmath}
\usepackage{multirow}
\usepackage{textcomp}
\usepackage{tabularx}
\usepackage{enumerate}
\usepackage{nameref}
\usepackage{graphicx}
\usepackage{appendix}
\usepackage{wrapfig}
\usepackage{makecell}

\usepackage{algorithm}
\usepackage{algpseudocode}
\usepackage{mathtools}

\usepackage{comment}
\newcommand*{\myfont}{\fontfamily{OldStandard}\selectfont}

\makeatletter
\def\RS#1{\zap@space#1 \@empty}
\makeatother

\usepackage{url}            
\usepackage{booktabs}       
\usepackage{amsfonts}       
\usepackage{nicefrac}       
\usepackage{microtype}      
\usepackage{xcolor}         

\title{Semi-Supervised Learning with Multi-Head~Co-Training}

\author{%
  Mingcai Chen$^1$, Yuntao Du$^1$, Yi Zhang$^1$, Shuwei Qian$^1$, Chongjun Wang$^1$\thanks{Corresponding authors}\\
}
\affiliations{
  $^1$State Key Laboratory for Novel Software Technology at Nanjing University\\
  Nanjing University, Nanjing 210023, China\\
  \texttt{\{chenmc, duyuntao, njuzhangy, QianSW\}@smail.nju.edu.cn}, \\
  \texttt{{chjwang}@nju.edu.cn} \\
}
\begin{document}
\maketitle
\begin{abstract}
    Co-training, extended from self-training, is one of the frameworks for semi-supervised learning.
    Without natural split of features, single-view co-training works at the cost of training extra classifiers, where the algorithm should be delicately designed to prevent individual classifiers from collapsing into each other.
    To remove these obstacles which deter the adoption of single-view co-training, we present a simple and efficient algorithm {\myfont Multi-Head\,Co-Training}. 
    By integrating base learners into a multi-head structure, the model is in a minimal amount of extra parameters.
    Every classification head in the unified model interacts with its peers through a “Weak and Strong Augmentation” strategy, in which the diversity is naturally brought by the strong data augmentation.
    Therefore, the proposed method facilitates single-view co-training by 1). promoting diversity implicitly and 2). only requiring a small extra computational overhead.
    The effectiveness of {\myfont Multi-Head\,Co-Training} is demonstrated in an empirical study on standard semi-supervised learning benchmarks.
\end{abstract}
\section{Introduction}
Benefiting from the rich data sources and growing computing power in the last decade, the field of machine learning has been thriving drastically. The advent of public datasets with a large amount of high-quality labels has further spawned many successful deep learning methods \cite{deng2009imagenet,he2016deep,zagoruyko2016wide,krizhevsky2012imagenet}. 
However, there could be various difficulties for obtaining label information, such as privacy, labor costs, safety or ethic issues, and requirement of domain experts \cite{zhou2018brief,10.5555/1841234,mahajan2018exploring}. 
All of these impel us to find a way of bringing unlabeled data into full play.
Semi-Supervised Learning (SSL) is a branch of machine learning which seeks to address the problem \cite{10.5555/1841234,chapelle2006continuation,prakash2014survey,van2020survey}. 
It utilizes both labeled and unlabeled data to improve performance. 

As one of the earliest and most popular SSL frameworks, self-training works by iteratively retraining the model using pseudo-labels obtained from itself \cite{lee2013pseudo,berthelot2019mixmatch,berthelot2019remixmatch,mclachlan1975iterative}.
Despite its simplicity and alignment to the task of interest \cite{zoph2020rethinking}, self-training underperforms due to the “confirmation bias” or “error accumulation”.
It means that some incorrect predictions could be selected as pseudo-labels to guide subsequent training, resulting in a loop of self-reinforcing errors \cite{zhang2016understanding}.

As an extension of self-training, co-training lets multiple individual models iteratively learn from each other \cite{zhou2010semi,wang2017theoretical}.
In the early multi-view co-training setting \cite{blum1998combining}, there should be a natural split of features where the “sufficiency and independence” assumptions should hold, i.e., it's sufficient to make predictions based on each view, and views are conditionally independent.
Later studies gradually reveal that co-training can also be successful in the single-view setting \cite{wang2017theoretical,dasgupta2002pac,abney2002bootstrapping,balcan2005co,wang2007analyzing}.  
Despite being feasible, the single-view co-training framework has received little attention recently. 
We attribute it to (a) the extra computational cost, which means at least twice model parameters of its self-training counterpart, and (b) the loss in simplicity, i.e., more design choices and hype-parameters are introduced for keeping individual classifiers uncorrelated. 

In this paper, we aim to facilitate the adoption of single-view co-training.
Inspired by recent developments of data augmentation and its applications in SSL \cite{berthelot2019mixmatch,sohn2020fixmatch,cubuk2020randaugment,devries2017improved}, 
we find that the enormous size of the augmentation search space naturally prevents base learners from converging to a consensus.
Employing the stochastic image augmentation frees us from delicately design different network structures or training algorithms.
Moreover, by replacing multiple individual models with a shared module followed by multiple classification heads, the model can achieve co-training in a minimal amount of extra parameters.
Combining these, we propose {\myfont Multi-Head\,Co-Training}, a new algorithm that facilitates the usage of single-view co-training. 
The main contributions are as follows:
\begin{itemize}
\item 
{\myfont Multi-Head\,Co-Training} addresses two obstacle factors of standard single-view co-training, i.e., extra design and computational cost.
\item Experimentally, we show that our method obtains state-of-the-art results on CIFAR, SVHN, and Mini-ImageNet.
Besides, We systematically study the components of {\myfont Multi-Head\,Co-Training}.
\item  We further analyze the calibration of SSL methods and provide insights regarding the link between confirmation bias and model calibration. 
\end{itemize}

\section{Related work} \label{relatedwork}
In this section, we concentrate on relevant studies to set the stage for {\myfont Multi-Head\,Co-Training}.
More extensive surveys on SSL can be found in \cite{prakash2014survey,van2020survey,zhu2005semi,zhou2010semi,zhou2018brief,subramanya2014graph}.

The basic assumptions in SSL are the smoothness assumption and low-density assumption. 
The smoothness assumption states that if two or more data points are close in the sample space, they should belong to the same class.
Similarly, the low-density assumption states that the decision boundary for a classification model shouldn't pass the high-density region of sample space. 
These assumptions are intuitive in vision tasks because an image with small noise is still semantically identical to the original one. 
A dominant paradigm in SSL is grounded on these assumptions. 
From this point of view, various ways of making use of unlabeled data, including consistency regularization, entropy minimization, perturbation-based methods, self-training, and co-training, are essentially similar.

Consistency regularization constrains the model to make consistent predictions across the same example under variants of noises. 
In a general form, 
\begin{equation}
D[q(y\mid x),p(y\mid x')]  
\end{equation}
where $q$ and $p$ are the modeled distributions. 
Different notations are used here, indicating that they could come from different models. 
The target example is denoted as $x$ and its noisy counterpart is denoted as $x'$. 
$D(\cdot, \cdot)$ can be any distance measurement, such as KL divergence or mean square error.
SSL methods falling into this category differ in the source of noise, models for two distributions, and the distance measurement. 
For example, VAT \cite{miyato2018virtual} generates noise in an adversarial direction. 
Laine \& Aila \cite{laine2016temporal} propose $\Pi$-Model and Temporal Ensembling. $\Pi$-Model performs Gaussian noise, dropout, etc., to augment images. 
Temporal Ensembling further ensembles prior network evaluations to encourage consistent predictions.
Mean Teacher \cite{tarvainen2017mean} instead maintains an Exponential Moving Average (EMA) of model's parameters.
ICT \cite{verma2019interpolation} applies consistency regularization between the prediction of interpolation of unlabeled points and the interpolation of the predictions at those points.
UDA \cite{xie2019unsupervised} replaces the traditional data augmentation with unsupervised data augmentation.

Self-training\footnote{In the field of SSL, the terminology “self-training” overlaps with “pseudo-labeling”, which refers to training the model incrementally instead of retraining in every iteration. For illustration, we use “self-training” throughout this paper, referring to a broad category of methods.}
favors low-density separation by using model's own predictions as pseudo-labels.
Pseudo-Labeling \cite{lee2013pseudo} picks the most confident predictions as hard (one-hot) pseudo-labels. 
Apart from that, the method also uses auto-encoder and dropout as regularization.
MixMatch \cite{berthelot2019mixmatch} uses the average of predictions on the image under multiple augmentations as the soft pseudo-label.
Furthermore, Mixup \cite{zhang2017mixup}, as a regularizer, is used to mix labeled and unlabeled data.
ReMixMatch \cite{berthelot2019remixmatch} improves MixMatch by introducing other regularization techniques and a modified version of AutoAugment \cite{cubuk2019autoaugment}.
FixMatch \cite{sohn2020fixmatch} finds an effective combination of image augmentation techniques and pseudo-labeling.
One of the reasons for the popularity of self-training is its simplicity.
It can be used with almost all supervised classifier \cite{van2020survey}.
Another important but rarely mentioned factor is its awareness of the task of interest.
Although some other unsupervised constraints may help build general representations, it has been shown that self-training can align to the task of interest well and benefit model in a more secure way \cite{zoph2020rethinking}.
However, confirmation bias, where the prediction mistakes would accumulate during the training process, damages the performance of self-training.

As an extension of self-training, co-training alleviates the problem of confirmation bias.
Two or more models are trained by each other's predictions.
In the original form \cite{blum1998combining}, two individual classifiers are trained on two views.
What's more, the proposed co-training algorithm requires the “sufficiency and independence” assumptions hold, i.e., two views should be sufficient to perform accurate prediction and be conditionally independent given the class.
Nevertheless, later studies show that weak independence \cite{abney2002bootstrapping,balcan2005co} or single-view data \cite{wang2017theoretical,wang2007analyzing,du2010does} is enough for a successful co-training algorithm.

Without the distinct views of the data containing complementary information, single-view co-training has to promote diversity in some other ways. 
It also has been shown that the more uncorrelated the individual classifiers are, the more effective the algorithm is. \cite{wang2017theoretical,wang2007analyzing}
Several early studies attempt to split the single-view data \cite{balcan2005co,chen2011automatic}, and some further approach a pure single-view setting by introducing diversity among the classifiers otherwise \cite{zhou2004democratic,goldman2000enhancing,xu2012dcpe}. 
Recently, Deep Co-training \cite{qiao2018deep} maintains disagreement through a view difference constraint. 
Tri-Net \cite{dong2018tri} adopts a multi-head structure. To prevent consensus, it designs different head structures and samples different sub-datasets for learners. 
CoMatch \cite{li2020comatch} applies a graph-based smoothness regularization and also integrates supervised contrastive learning.
Although have achieved a number of successes, these methods complicate the practical adoption of co-training. 
In {\myfont Multi-Head\,Co-Training}, employing the stochastic image augmentation frees us from delicately designing different network structures or training algorithms. Unique to other methods, no preventive measures, e.g., extra loss term or different base learners, are needed to avoid collapse. Along with the reduction of computational cost brought by the multi-head structure, the proposed method facilitates the adoption of single-view co-training.

\section{Multi-Head~Co-Training} \label{method}

\begin{figure}[t]
    \centering
    \includegraphics[width=1\columnwidth]{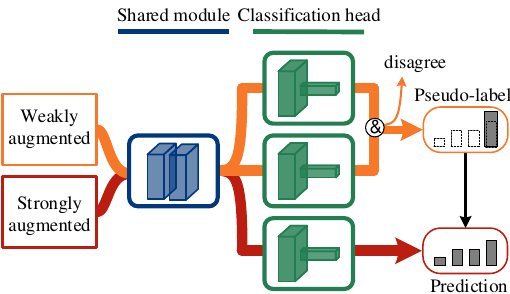} 
    \caption{Diagram of {\myfont Multi-Head\,Co-Training} with three heads.
    Images are fed into a shared module (blue box) followed by three classification heads (green boxes). 
    Among them, weakly augmented images (orange lines) are for pseudo-labeling.
    The pseudo-labels are to guide the predictions on strongly augmented examples (red line).
    Here, pseudo-labels for the bottom head are generated and selected according to the other two heads’ predicted classes on the weakly augmentation images.
    \textit{Note that only the co-training process of the bottom head is shown here.}
    The weakly and strongly augmented images are in fact simultaneously fed into all three heads.
    }
    \label{MM}
\end{figure}

In this section, we introduce the details of {\myfont Multi-Head\,Co-Training}.
Formally, for a $C$-class classification problem, SSL aims to model a class distribution $p(y\mid x)$ for input $x$ utilizing both labeled and unlabeled data.
In {\myfont Multi-Head\,Co-Training}, the parametric model consists of a shared module $f$ and multiple classification heads $g_m$ ($m\in\{1,\dots,M\}$) with the same structure. 
Let $p_m(y\mid x)=g_m(f(x))$ represent the predicted class distribution produced by $f$ and $g_m$.
All classification heads are updated using the using a consensus of predictions from other heads. 
Apart from that, following the principle of recent successful SSL algorithms, this method utilizes image augmentation techniques to employ a “Weak and Strong Augmentation” strategy. 
It uses predictions on weakly augmented images, which are relatively more accurate, to correct the predictions on strongly augmented images.
The weak and strong augmentation function are denoted as $Aug_w(\cdot)$ and $Aug_s(\cdot)$ respectively and further introduced in \nameref{dataaug}.
The diagram of {\myfont Multi-Head\,Co-Training} with three heads is shown in Figure~\ref{MM}.
The overall algorithm is shown in Algorithm \ref{algorithm}.

\begin{algorithm}
    \caption{{\myfont Multi-Head\,Co-Training} with three heads.} 
    \label{algorithm}
    \textbf{Input}: Labeled batch $\mathcal{D}_l={\{(x_b,y_b)\};b\in {1,\dots,B_l}}$, unlabeled batch $\mathcal{D}_u={\{(u_b);b\in {1,\dots,B_u}\}}$, unsupervised loss weight $\lambda$. \\
    \textbf{Output}: Updated model.
    \begin{algorithmic}[1] 
    \For{$b=1$ to $B_l$}
        \State $\hat{x}_b = Aug_w(x_b)$ 
        \Comment{weakly augment} 
    \EndFor
    \State $\mathcal{L}_{l} =\frac{1}{B_l}\sum^{B_l}_{b=1}\sum_{m\in\{1,2,3\}} \text{H}(y_b,p_m(y\mid \hat{x}_b))$ 
    \For{$b=1$ to $B_u$}
        \State $\hat{u}_b=Aug_w(u_b)$ 
        \Comment{weakly augment} 
        \For{$m=1$ to 3}\Comment{pseudo-labeling}
            \State $q_{b,m} = \mathop{\arg\max}_{c} p_i(y\mid \hat{u}_b) \quad i\neq m $  
        \EndFor
        \For{$m=1$ to 3}\Comment{selected agreed examples}
            \State $\mathds{1}_{b,m} =[{q_{b,i}=q_{b,j}}] \quad \{i,j\}=\{1,2,3\} \setminus m$ 
        \EndFor
        \For{$m=1$ to 3} \Comment{unsupervised loss}
            \State $\tilde{u}_{b,m}=Aug_s(u_b)$ 
            \Comment{strongly augment} 
            \State $\ell_{b,m}=\text{H}(q_{b,i},p_m(y\mid \tilde{u}_{b,m})) \quad i\neq m$ 
        \EndFor
    \EndFor
    \State $\mathcal{L}_{u} =\frac{1}{3}\sum_{m\in\{1,2,3\}}\frac{1}{\sum_{b=1}^{B_u}\mathds{1}_{b,m}}\sum_{b=1}^{B_u}\mathds{1}_{b,m} \ell_{b,m} $ 
    \State update all parameters according to $\mathcal{L}_{l}+\lambda\mathcal{L}_{u}$
    \end{algorithmic}
\end{algorithm}

\subsection{The multi-head structure} \label{mmstructure}
In every training iteration of {\myfont Multi-Head\,Co-Training}, a batch of labeled examples $\mathcal{D}_l={\{(x_b,y_b);b\in (1,\dots,B_l)\}}$ and a batch of unlabeled examples $\mathcal{D}_u=\{(u_b);b\in (1,\dots,B_u)\}$ are randomly sampled from the labeled and unlabeled dataset respectively.
For supervised training, the parameters in the shared module $f$ and all heads $g_m$ ($m\in \{1,\dots,M\}$) are updated to minimize the cross-entropy loss between predictions and true labels, i.e., 
\begin{equation}
    \mathcal{L}_{l} =\frac{1}{B_l}\sum^{B_l}_{b=1}\sum_{m=1}^{M} \text{H}(y_b,p_m(y\mid \hat{x}_b))\\
\end{equation}
where $\text{H}(\cdot,\cdot)$ represents the cross-entropy loss function and $\hat{x}_b=Aug_w(x_b)$ is the weakly augmented labeled example. 

For co-training, every head interacts with its peers through pseudo-labels on unlabeled data.
To obtain reliable predictions for pseudo-labeling, weakly augmented unlabeled examples $\hat{u}_b=Aug_w(u_b)$ first pass through the shared module and all heads simultaneously,
\begin{equation}
    q_{b,m} = \mathop{\arg\max}_c p_m(y=c\mid \hat{u}_b)\\
\end{equation}
where $\mathop{\arg\max}$ picks the class $c\in \{1,\dots,C\}$ with the maximal probability, i.e.,  the most confident predicted class $q_{b,m}$.
To avoid confirmation bias, pseudo-labels for each head depend on the predicted classes from other $M-1$ heads,
\begin{equation}
    \mathds{Q}_{b,m}=\{q_{b,1},\dots,q_{b,M}\} \setminus q_{b,m}
\end{equation}
where $\setminus$ is the set operation of removing element.
The most frequently predicted class in multiset $\mathds{Q}_{b,m}$ is the pseudo-label $\bar{q}_{b,m}$ for $m$-th head, and it is selected only when more than half heads agree on the prediction, 
\begin{equation}
    \begin{aligned}
        \bar{q}_{b,m} &= \mathop{\arg\max}_{c}\sum_{q\in\mathds{Q}_{b,m}}[q=c] \\
        \mathds{1}_{b,m} &=[\sum_{q\in\mathds{Q}_{b,m}} [q=\bar{q}_{b,m}] > \frac{M}{2}]
    \end{aligned}
\end{equation}
where $[\cdot]$ refers to the Iverson brackets, defined to be 1 if the statement in it is true, otherwise 0. 
$\mathds{1}_{b,m}$ indicates whether $b$-th pseudo-label is selected for $m$-th head.
After the selection process, uncertain (less agreed) examples are filtered.
In the meantime, strongly augmented unlabeled examples go through the shared module $f$ and corresponding head $g_m$.
The average cross-entropy is then calculated:
    
\begin{equation}
\begin{aligned}
    \mathcal{L}_{u} &=\frac{1}{M}\sum_{m=1}^{M}\frac{1}{\sum_{b=1}^{B_u}\mathds{1}_{b,m}}\sum_{b=1}^{B_u}\mathds{1}_{b,m}\ell_{b,m}  \\
    \ell_{b,m}&=\text{H}(\bar{q}_{b,m},p_m(y\mid \tilde{u}_{b,m}))
\end{aligned}
\end{equation}
where $\{(\tilde{u}_{b,m});m\in (1,\dots,M)\}$ comes from $M$ times of strong augmentation $Aug_s(\cdot)$.
Note that the transformation function generates a differently augmented image every time.
The supervised loss and unsupervised loss are added together as the total loss $\mathcal{L}$ (weighted by coefficient $\lambda$), i.e.,
\begin{equation}
    \mathcal{L}=\mathcal{L}_{l}+\lambda\mathcal{L}_{u}
\end{equation}
The algorithm proceeds until reaching fixed iterations (training details are illustrated in the supplementary material).

As discussed in \nameref{relatedwork}, the quality of pseudo-labels and the diversity between individual classifiers are the two important things for a co-training algorithm to succeed.
In our method, the diversity between heads inherently comes from the randomness in the strong augmentation function (consequently, the unlabeled examples for each head are differently augmented, selected, and pseudo-labeled).
Unique to co-training algorithms, diversity is promoted implicitly in {\myfont Multi-Head\,Co-Training}.
In terms of the quality of pseudo-labels, ensemble predictions of other heads on weakly augmented examples are used for accurately pseudo-labeling and selecting.

\subsection{The “Weak and Strong Augmentation” strategy} \label{dataaug}
Due to the scarcity of labels, preventing overfitting, or in other words, improving the generalization ability is the core task of SSL.
Data augmentation approaches such a problem via expanding the size of training set, and thus plays a vital role in SSL.
In {\myfont Multi-Head\,Co-Training}, the weak and strong augmentation functions differ in the degree of image augmentation.
Specifically, the weak image transformation function $Aug_w(\cdot)$ applies random horizontal flip and random crop. 
Two augmentation techniques, namely RandAugment \cite{cubuk2020randaugment} and Cutout \cite{devries2017improved}, constitute the strong transformation function $Aug_s(\cdot)$. 
In RandAugment, a given number of operations are randomly selected from a fixed set of geometric and photometric transformations, such as affine transformation, color adjustment.
Then they are applied to images with a given magnitude.
Cutout randomly masks out square regions of images.
Both of them are applied sequentially in the strong augmentation. 
It has been shown that unsupervised learning benefits from stronger data augmentation \cite{chen2020simple}. 
The same preference can also be extended to SSL.
Thus, the setting of RandAugment follows the modified stronger version in FixMatch \cite{sohn2020fixmatch} and details are reported in the supplementary material.
    


\subsection{Exponential moving average} 
To enforce smoothness, Exponential Moving Average (EMA) is a widely used technique in SSL. 
In this paper, we maintain an EMA model for evaluation. Its parameters are updated every iteration using the training-time model's parameters:
\begin{equation}
    \begin{aligned}
        \overline{\theta}\leftarrow\alpha\overline{\theta}+(1-\alpha)\theta
    \end{aligned}
\end{equation}
where $\overline{\theta}$ is the parameters of the EMA model.
$\theta$ is the parameters of training-time model. 
$\alpha$ is the decay rate which controls how much the average model moves every time.
During test-time, we simply ensemble all heads' predictions of the EMA model by adding them together.



\section{Experiments} \label{experiments}
The number and structure of heads in our framework can be arbitrary, but we set the head number as three and the head structure as the last residual block \cite{zagoruyko2016wide} in most of the experiments. 
The choice is the result of a trade-off between accuracy and efficiency (illustrated in \nameref{Ablation Study}).

\subsection{Results} \label{results}

\begin{table*}
    \centering
    \begin{tabular}{lrrrrrrr}
    \toprule
                              & \multicolumn{3}{c}{SVHN}&\multicolumn{3}{c}{CIFAR-10}& CIFAR-100 \\  \cmidrule{1-1} \cmidrule(l){2-4} \cmidrule(l){5-7} \cmidrule(l){8-8} 
    Method                 & 250 labels            &      500 labels       &     1000 labels           & 250 labels                 & 1000 labels          & 4000 labels            & 10000 labels           \\ \midrule                                                                                                              %
    Tri-net                &  -\hspace{7mm}        &   -\hspace{7mm}       &    \RS{ 3.71 $\pm$ 0.14}  &  -\hspace{7mm}             & -\hspace{7mm}        & \RS{8.45$\pm$0.22}     &  -\hspace{7mm}         \\ 
    $\Pi$-Model            &  \RS{17.65$\pm$0.27}  &   \RS{11.44$\pm$0.39} &    \RS{ 8.60 $\pm$ 0.18}  & \RS{53.02$\pm$2.05}        & \RS{31.53$\pm$0.98}  & \RS{17.41$\pm$0.37}    &  \RS{37.88$\pm$0.11}   \\ 
    Pseudo-Label           &  \RS{21.16$\pm$0.88}  &   \RS{14.35$\pm$0.37} &    \RS{10.19 $\pm$ 0.41}  & \RS{49.98$\pm$1.17}        & \RS{30.91$\pm$1.73}  & \RS{16.21$\pm$0.11}    &  \RS{36.21$\pm$0.19}   \\  
    VAT                    &  \RS{ 8.41$\pm$1.01}  &   \RS{ 7.44$\pm$0.79} &    \RS{ 5.98 $\pm$ 0.21}  & \RS{36.03$\pm$2.82}        & \RS{18.68$\pm$0.40}  & \RS{11.05$\pm$0.31}    &  -\hspace{7mm}         \\  
    Mean Teacher           &  \RS{ 6.45$\pm$2.43}  &   \RS{ 3.82$\pm$0.17} &    \RS{ 3.75 $\pm$ 0.10}  & \RS{47.32$\pm$4.71}        & \RS{17.32$\pm$4.00}  & \RS{10.36$\pm$0.25}    &  \RS{35.83$\pm$0.24}   \\  
    MixMatch               &  \RS{ 3.78$\pm$0.26}  &   \RS{ 3.27$\pm$0.31} &    \RS{ 3.27 $\pm$ 0.31}  & \RS{11.08$\pm$0.87}        & \RS{ 7.75$\pm$0.32}  & \RS{ 6.24$\pm$0.06}    &  \RS{28.31$\pm$0.33}   \\  
    ReMixMatch             &  \RS{ 3.10$\pm$0.50}  &  -\hspace{7mm}        &    \RS{ 2.83 $\pm$ 0.30}  & \RS{ 6.27$\pm$0.34}        & \RS{ 5.73$\pm$0.16}  & \RS{ 5.14$\pm$0.04}    &  \RS{23.03$\pm$0.56}   \\  
    FixMatch (RA)          &  \RS{ 2.48$\pm$0.38}  &   -\hspace{7mm}       &    \RS{ 2.28 $\pm$ 0.11}  & \RS{ 5.07$\pm$0.65}        & -\hspace{7mm}        &\RS{ 4.26$\pm$0.05}     &  \RS{22.60$\pm$0.12}   \\
    FixMatch (CTA)         &  \RS{ 2.64$\pm$0.64}  &   -\hspace{7mm}       &    \RS{ 2.36 $\pm$ 0.19}  & \RS{ 5.07$\pm$0.33}        & -\hspace{7mm}        & \RS{ 4.31$\pm$0.15}    &  \RS{23.18$\pm$0.11}   \\
    Ours                   &\textbf{2.21$\pm$0.18} &\textbf{2.18$\pm$0.08} &    \textbf{2.16$\pm$0.05} &\textbf{4.98$\pm$0.30}      &\textbf{4.74$\pm$0.16}&\textbf{3.84$\pm$0.09}  &\textbf{21.68$\pm$0.16} \\

    \bottomrule
    \end{tabular}
    \caption{Error rates for SVHN, CIFAR-10, and CIFAR-100. The best results are in bold.}
    \label{all results}
\end{table*}

We benchmark the proposed method on experimental settings using CIFAR-10 \cite{krizhevsky2009learning}, CIFAR-100 \cite{krizhevsky2009learning}, and SVHN \cite{netzer2011reading} as the standard practice.
Different portions of labeled data ranging from 0.5\% to 20\% are experimented.
For comparison, we consider Tri-net \cite{dong2018tri}, $\Pi$-Model \cite{laine2016temporal}, Pseudo-Label \cite{lee2013pseudo}, Mean Teacher \cite{tarvainen2017mean}, VAT \cite{miyato2018virtual}, MixMatch \cite{berthelot2019mixmatch}, ReMixMatch \cite{berthelot2019remixmatch}, FixMatch \cite{sohn2020fixmatch}.
The results of these methods reported in this section are reproduced by \cite{berthelot2019mixmatch,berthelot2019remixmatch} using the same backbone and training protocol (except for the results of Tri-net and FixMatch are from their papers).
The main criterion is the error rate. 
Variance is also reported to ensure the results are statistically significant (5 runs with different seeds).
We report the final performance of the EMA model. 
SGD with Nesterov momentum \cite{sutskever2013importance} is used, along with weight decay and cosine learning rate decay \cite{loshchilov2016sgdr}.
The details are in the supplementary material.
\subsubsection{CIFAR-10, CIFAR-100} \label{cifar}
We first compare {\myfont Multi-Head\,Co-Training} to the state-of-the-art methods on CIFAR \cite{krizhevsky2009learning}, which is one of the most commonly used image recognition datasets. 
We randomly choose 250-4000 from 50000 training images of CIFAR-10 as labeled examples.
Other images' labels are thrown away. 
The backbone model is WRN 28-2 (extra heads are added).
As shown in Table~\ref{all results}, {\myfont Multi-Head\,Co-Training} performs consistently better against the state-of-the-art methods.
For example, it achieves an average error rate of 3.84\% on CIFAR-10 with 4000 labeled images, compared favorably to the state-of-the-art results.

We randomly choose 10000 from 50000 training images of CIFAR-100 as labeled examples and throw other's label information. 
In Table~\ref{all results}, we present the results of CIFAR-100 with 10000 labels. 
As the common practice in recent methods, WRN~28-8 is used to accommodate the more challenging task (more classes and each with fewer examples). 
We reduce the number of channels of the final block in WRN~28-8 from 512 to 256.
By doing so, the model has a much smaller size.
Combining the results in Table~\ref{all results}, {\myfont Multi-Head\,Co-Training} achieves 21.68\% error rate, having an improvement of 1.5\% compared to the best results of previous methods with even a smaller model.

\begin{table}
    \centering
    \begin{tabular}{lcc} \toprule
                    & 4000 labels                   &  10000 labels\\ \midrule 
        Mean Teacher& \RS{72.51$\pm$0.22}           & \RS{57.55$\pm$1.11}\\
        Pseudo-Label& \RS{56.49$\pm$0.51}           & \RS{46.08$\pm$0.11}\\
        LaplaceNet  & \textbf{46.32$\pm$0.27}           & \textbf{39.43$\pm$0.09} \\
        Ours        & \RS{46.53$\pm$0.15}           & \RS{39.74$\pm$0.12} \\
        \bottomrule
    \end{tabular}
    \caption{Error rates for Mini-ImageNet.}
    \label{params}
\end{table}

\subsubsection{SVHN}

Similarly, we evaluate the accuracy of our method with a varying number of labels from 250 to 1000 on the SVHN dataset \cite{netzer2011reading} (the extra training set is not used). 
The image augmentation for SVHN is different because some operations are not suitable for digit images (e.g., horizontal flip for asymmetrical digit images).
Its details are in the supplementary material.
The results of {\myfont Multi-Head\,Co-Training} and other methods are shown in Table~\ref{all results}.
{\myfont Multi-Head\,Co-Training} outperforms other methods by a small margin.

\subsubsection{Mini-ImageNet}

We further evaluate our model on the more complex dataset Mini-ImageNet \cite{vinyals2016matching}, which is a subset ImageNet \cite{deng2009imagenet}.
The training set of Mini-ImageNet consists of 50000 images with a size of 84 × 84 in 10 object classes. 
We randomly choose 4000 and 10000 images as labeled examples and throw other's label information. 
The backbone model is ResNet-18 \cite{wang2017residual} and early stopped using the ImageNet validation set.
Other methods' results are from \cite{sellars2021laplacenet}.
Our method achieves an error rate of  47.88\% and 39.74\% for 4k and 10k labeled images, respectively. 
The results are competitive to a recent method LaplaceNet \cite{sellars2021laplacenet}, which uses graph-based constrain and multiple strong augmentation.
Besides, our co-training method, which is simple and efficient, is orthogonal to other SSL constraints.


\begin{table}
    \centering
    \begin{tabular}{lcc} \toprule
                    & WRN 28-2 & WRN 28-8 \\ \midrule 
        Original    & 1.4 M (30 min)   & 23.4 M (136 min) \\
        Three model & 4.2 M (66 min)   & 70.2 M (344 min) \\
        Ours        & 3.7 M (39 min)   & 19.9 M (168 min)  \\
        \bottomrule
    \end{tabular}
    \caption{Number of parameters (Million) and average training time for 10000 iterations (minutes).}
    \label{params}
\end{table}

\subsubsection{Computational cost analysis}
We report the training cost of the original backbone, standard co-training with three models, and our method in Table~\ref{params}. The reduction of the number of parameters and training time is significant. For example, the number of parameters in the WRN 28-8 backbone is instead fewer than the original one. Only 23.5\% extra time, compared to self-training, is needed to train our method, while the standard co-training needs 152.9\% extra time.

\subsection{Ablation study} \label{Ablation Study}
This section presents an ablation study to measure the contribution of different components of {\myfont Multi-Head\,Co-Training}.
Specifically, we measure the effect of 
\begin{enumerate}[1).]
\item \label{onehead}{\myfont Multi-Head\,Co-Training} with only one head. Pseudo-labels are generated from their own predictions and selected by a confidence threshold of 0.95. 
\item \label{onestrong}{\myfont Multi-Head\,Co-Training} with one strong augmentation. Strong augmentation is only performed once and forwarded to all three heads.
\item \label{noweak}{\myfont Multi-Head\,Co-Training} without weak augmentation. The original images are for pseudo-labeling.
\item \label{sameini}{\myfont Multi-Head\,Co-Training} with three heads with the same initialization.
\item {\myfont Multi-Head\,Co-Training} without EMA.
\end{enumerate}

\begin{table}
    \centering
    \begin{tabular}{lcc} 
        \toprule
        Ablation & One head & Ensemble \\ \midrule 
        {\myfont Multi-Head\,Co-Training} &  4.22  &  3.84   \\   
        1). One head                                               &  4.43  &  4.23   \\   
        2). One strong augmentation                                &  4.45  &  4.03   \\   
        3). W/o weak augmentation                                  &  4.86 &   4.55 \\   
        4). Same heads' initialization                             &  4.28  &  3.86   \\  
        5). W/o EMA                                                   &  6.23  &  5.30   \\ \bottomrule 
    \end{tabular}
\caption{Ablation experiments. The models are trained on CIFAR-10 with 4000 labels. The average error rate of individual heads and their ensemble both reported.}
\label{Ablation}
\end{table}

We first set {\myfont Multi-Head\,Co-Training}'s self-training counterpart as described at \ref{onehead}) as a baseline. 
It has the same backbone and hype-parameters but with only one head.
The self-training baseline obtains sub-optimal performance.
To further verify the main improvement of our method is not coming from ensembling, the self-training model is retrained three times with different initialization to produce an ensemble result (``Ensemble'' in \ref{onehead}) row).
It can be observed from Table~\ref{Ablation} that {\myfont Multi-Head\,Co-Training}, as a co-training algorithm, first shows its effectiveness by outperforming it.


Promoting diversity between individual models is critical to the success of the co-training framework. 
Otherwise, they would produce too similar predictions, and thus co-training degenerates into self-training. 
Other single-view co-training methods create diversity mainly in several ways, including automatic view splitting \cite{balcan2005co,chen2011automatic}, using different individual classifiers or individual classifiers with different structures \cite{zhou2004democratic,goldman2000enhancing,xu2012dcpe,dong2018tri}. 
Unique to a co-training algorithm, Multi-Head~Co-Training doesn't promote diversity explicitly.
The diversity between heads inherently comes from the randomness in parameter initialization and augmentation (consequently, examples selected for each head are different).
To study the impact of them, we remove each source of diversity at \ref{onestrong}) and \ref{sameini}) respectively.
As shown in Table~\ref{Ablation}, the accuracy drops more when differently augmented images are missing.
Moreover, the individual heads' error rate is almost the same as the self-training baseline \ref{onehead}).
This shows the important role strong augmentation plays in {\myfont Multi-Head\,Co-Training}.
As a regularizer, data augmentation is considered to confine learning to a subset of the hypothesis space \cite{zhang2016understanding}.
We believe multiple strong augmentations confine classification heads of {\myfont Multi-Head\,Co-Training} in different subsets of the hypothesis space and thus, keeping them uncorrelated.

\begin{figure}
        \centering
        \subfloat[Different number of heads.]{%
            \includegraphics[width=0.95\columnwidth]{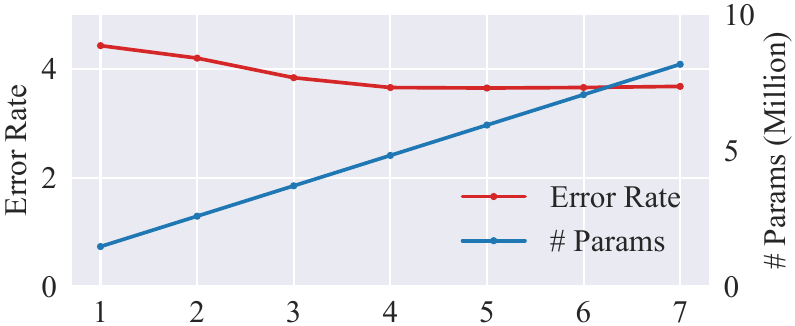}
            \label{fig_numheads}
        }\\
            \subfloat[Different structure of heads.]{%
            \includegraphics[width=0.95\columnwidth]{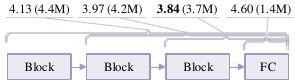}
            \label{fig_share}
            }
        \caption{The error rate and the number of parameters brought by different heads.}
\end{figure}

According to the observation in \ref{noweak}), replacing the weakly augmented images with the original ones leads to a worse final performance.
Note that pseudo-labels obtained from original images are more accurate.  
It means that current pseudo-labels from weakly augmented images, even with lower accuracy, would lead to a better model in later training.
An interesting fact implied by the phenomenon is that accuracy isn't the only important factor of pseudo-labels.


\subsubsection{The number and structure of heads} \label{numheads}

One main difference between {\myfont Multi-Head\,Co-Training} and other SSL methods is the multi-head structure. 
It brings many benefits. 
Firstly, it naturally produces multiple uncorrelated predictions for each example that regularizes the feature representation the shared module learns.
Secondly, pseudo-labels coming from the ensemble predictions on weakly augmented examples are more reliable. 
Thirdly, the number of parameters is much smaller because base learners share a module.
Based on WRN \cite{zagoruyko2016wide}, we empirically find a both accurate and efficient structure.
Specifically, we experiment {\myfont Multi-Head\,Co-Training} with different number of heads and different head structures in this section.
Considering that it's impractical to search all combinations, WRN 28-2 backbone on CIFAR-10 with 4000 labels is studied.
The head structure is fixed when we attempt to find the optimal number of heads.
Similarly, the number of heads is fixed when we attempt to find the optimal head structure.

We first test {\myfont Multi-Head\,Co-Training} with 1-7 heads while fixing the structure of head as the last block in WRN 28-2.
For consistency, when the number of heads is 1 or 2, i.e., the pseudo-labels come from only one head's predictions, a threshold of 0.95 is used for selecting.
As shown in Figure~\ref{fig_numheads}, with more heads, better performance can be obtained, but the accuracy growth slows down.
The structures with more heads are not considered because the gain becomes insignificant with the increasing of the number of heads.

\begin{figure*}
        \centering
        \subfloat[FixMatch]{%
            \includegraphics[width=0.235\textwidth]{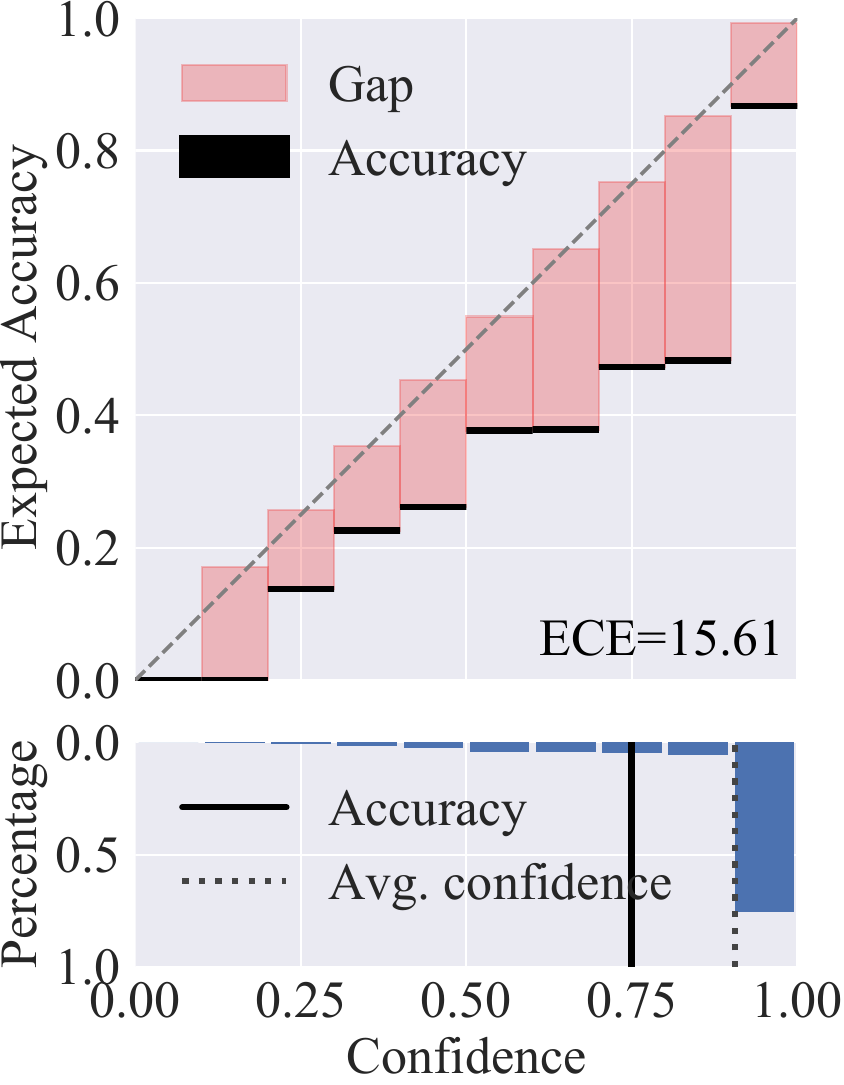}
            \label{cal_FM}
        }
        \subfloat[Ours]{%
            \includegraphics[width=0.235\textwidth]{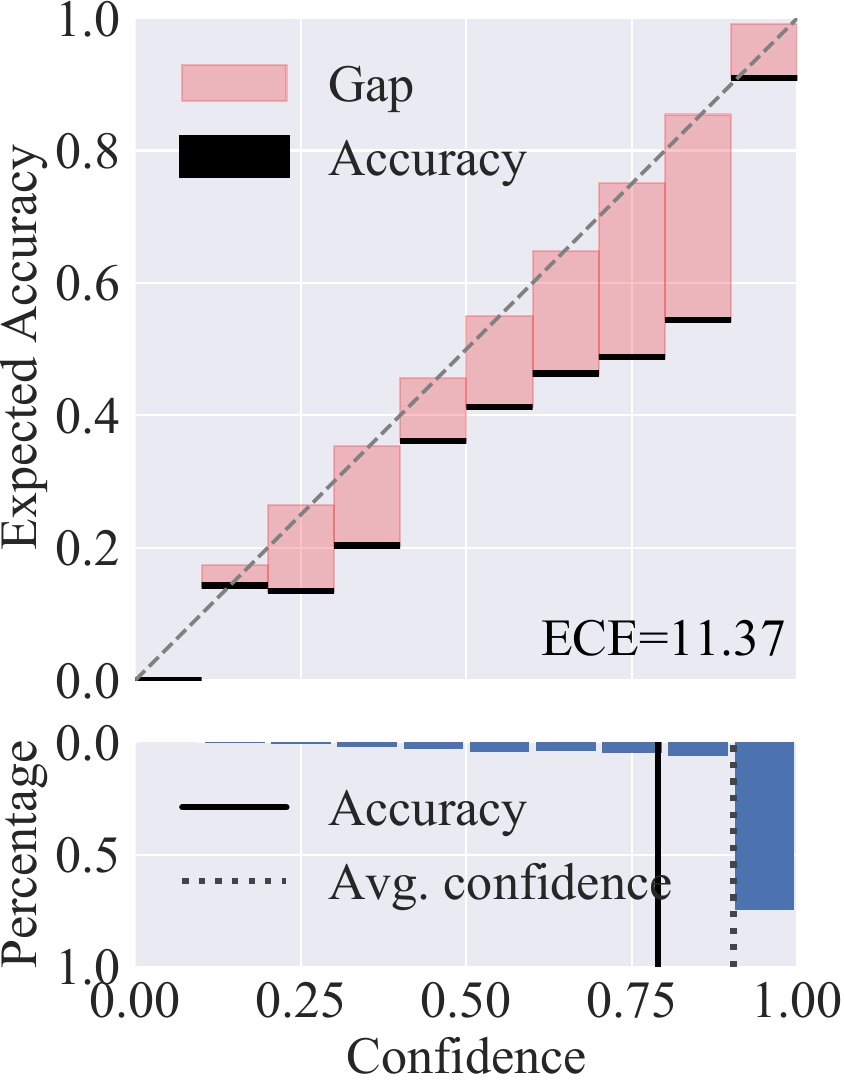}
            \label{cal_all}
        } 
        \subfloat[Calibrated FixMatch]{%
            \includegraphics[width=0.235\textwidth]{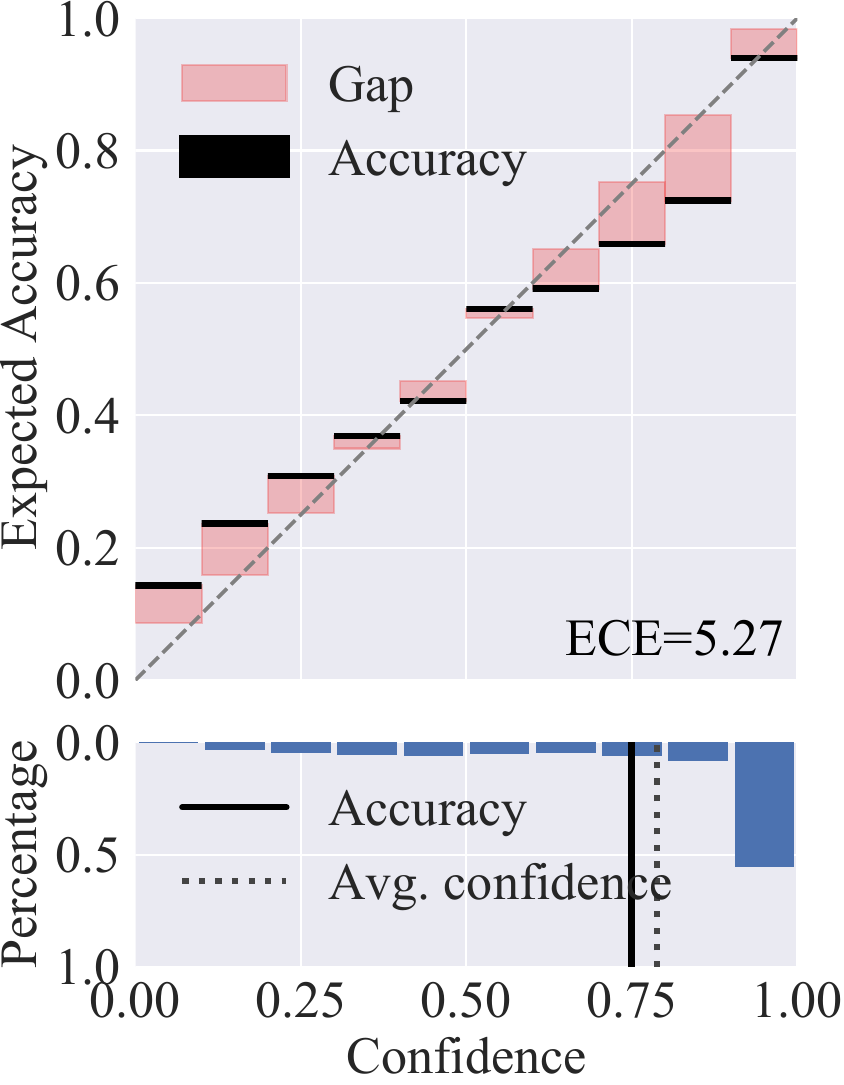}
            \label{cal_FM_s}
        }
        \subfloat[Calibrated ours]{%
            \includegraphics[width=0.235\textwidth]{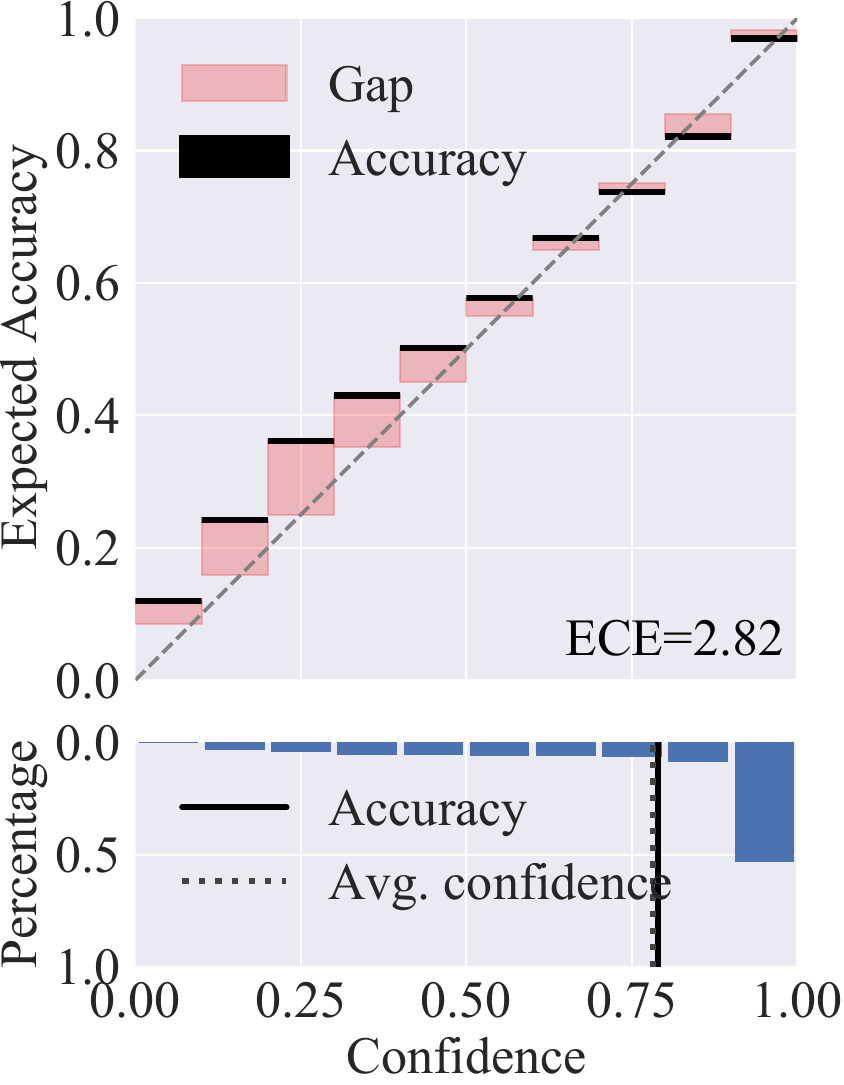}
            \label{cal_all_s}
        } 
        \caption{Reliability diagrams (top) and confidence histograms (bottom).\protect\footnotemark\,The models are trained on CIFAR-100 with 10000 labels, and their predictions on the test set are grouped into 10 interval bins (horizontal axis). Reliability diagram presents the true accuracy, the expected accuracy, and the gap between them of each bin. confidence histogram presents the percentage of examples that falls into each bin. The accuracy and average confidence are indicated by the solid and dashed lines, respectively.} 
        \label{cal}
\end{figure*}
 
In terms of the head structure, the most important thing is finding the balance point between the size of the shared module and the size of the head.
As shown in Figure~\ref{fig_share}, the best performance is observed when the heads have the structure of one block and one fully connected layer and the shared module has the structure of one convolutional layer and two blocks.
Whether increase or decrease the size of the heads would damage performance. 
Our explanation is that if there are too many shared parameters, there could be little room for the heads to make diverse predictions.
If there are too many independent parameters, the heads would easily fit the pseudo-labels, and again, fail to make diverse predictions.

Considering our main purpose of developing an effective co-training algorithm, we set the number of heads as three and the head as one block in our experiments.



\section{Calibration of SSL} \label{Calibration of SSL}
Confirmation bias comes up frequently in SSL literature.
However, it is hard to formulate or observe the problem.
We notice that most self-training or co-training methods select pseudo-labels by some criteria, such as confidence threshold.
In these cases, confirmation bias is closely related to the over-confidence of the network:
Wrong predictions with high confidence are likely to be selected and then used as pseudo-labels.
Thus, we link confirmation bias to model calibration, i.e., the problem of predicted probability represents the true correctness likelihood. 
We envision the calibration measurements be used to evaluate confirmation bias and help the design of self-training and co-training algorithms.
Apart from this, the challenges of SSL and calibration could appear simultaneously in real-world.
For example, in one of the applications of SSL, medical diagnosis, control should be passed on to human experts when the confidence of automatic diagnosis is low. 
In such scenarios, a well-calibrated SSL model is needed. 
To the best of our knowledge, these two problems have hitherto been studied independently. 
\footnotetext{Draw with the code in \url{https://github.com/hollance/reliability-diagrams}.}

According to our observation, SSL models have poor performance in terms of calibration due to entropy minimization and other similar constraints.
We analyze FixMatch (implemented using the same code-base), as one of the typical SSL methods, and {\myfont Multi-Head\,Co-Training} on CIFAR-100 with 10000 labels from the perspective of model calibration.
Several common calibration indicators are used, namely Expected Calibration Error (ECE), confidence histogram, and reliability diagram (illustrated in the supplementary material).
As shown in Figure~\ref{cal_FM}, FixMatch  has an average confidence of 78.75\% but only 74.93\% accuracy, producing over-confident results with an ECE value of 15.61. 
In Figure~\ref{cal_all}, we show the results of {\myfont Multi-Head\,Co-Training}.
Although our method produces over-confident predictions, the ECE value is smaller, indicating better probability estimation.

To further investigate, we apply a simple calibration technique called “temperature scaling” \cite{guo2017calibration} (see the supplementary material). 
From the calibrated results in Figure~\ref{cal_FM_s} and Figure~\ref{cal_all_s}, it can be observed that the miscalibration is remedied. 
The ECE value of calibrated FixMatch is improved to 5.27\% 
{\myfont Multi-Head\,Co-Training}'s reliability diagrams closely recovers the desired diagonal function with a low ECE value of 2.82.
It can be concluded that {\myfont Multi-Head\,Co-Training} produces good probability estimates naturally and can be better calibrated with simple techniques. 
Therefore, we suggest that confirmation bias is better addressed in our method from the perspective of calibration.

\section{Conclusion} \label{conclusion}
The field of SSL encompasses a broad spectrum of algorithms. 
However, co-training framework has received little attention recently because of the diversity criterion, extra computational cost.
{\myfont Multi-Head\,Co-Training} pointedly addresses these. 
It achieves single-view co-training by integrating the individual models into one multi-head structure and utilizing the data augmentation techniques.
As a result, the proposed method is 1). in a minimal amount of both parameters and hype-parameters and 2). doesn't need extra effort to promote diversity.
We present systematic experiments to show that {\myfont Multi-Head\,Co-Training} is a successful co-training method and outperforms state-of-the-art methods.
The solid empirical results suggest that it is possible to scale co-training to more realistic SSL settings.
In future work, we are interested in combining modality-agnostic data augmentation to make {\myfont Multi-Head\,Co-Training} ready to be applied to other tasks.


\clearpage
\section{Acknowledgements}
This paper is supported by the National Key Research and Development Program of China (Grant No. 2018YFB1403400), the National Natural Science Foundation of China (Grant No. 61876080), the Key Research and Development Program of Jiangsu(Grant No. BE2019105), the Collaborative Innovation Center of Novel Software Technology and Industrialization at Nanjing University.

\bibliography{ref}


\clearpage
\appendix
\begin{appendices}

\section{Notation and definitions}

\begin{table}[H]
    \centering
    \begin{tabularx}{\columnwidth}{lX}
        \toprule
        Notation            & Definition \\ \midrule
        $\text{H}(\cdot,\cdot)$          & Cross-entropy between two distributions. \\\midrule
        $\mathcal{D}$       & A batch of examples. $\mathcal{D}_l$ and $\mathcal{D}_u$ are a batch of labeled examples and unlabeled examples respectively.\\\midrule
        $p$                 & Modeled distribution. $p_m$ corresponds to $m$-th head.\\\midrule
        $B$                 & Batch size. $B_l$ and $B_u$ is the number of labeled and unlabeled examples respectively.\\\midrule
        $x$                 & Labeled example. \\\midrule
        $y$                 & Label. \\\midrule
        $u$                 & Unlabeled example.\\\midrule
        $q$                 & Predicted class. $q_{b,m}$ is the predicted class of $m$-th head on $b$-th example. $\bar{q}_{b,m}$ is the pseudo-label for $m$-th head on $b$-th example.\\\midrule
        $Aug(\cdot)$        & Augmentation function. $Aug_w(\cdot)$ and $Aug_s(\cdot)$ are the weak augmentation function and strong augmentation function respectively. \\\midrule
        $f$                 & Shared module in {\myfont Multi-Head\,Co-Training}. \\\midrule
        $g$                 & Classification head in {\myfont Multi-Head\,Co-Training}. $g_i$ is the $m$-th head. \\\midrule
        $[\cdot]$           & The Iverson brackets, defined to be 1 if the statement in it is true, otherwise 0.\\\midrule
        $\mathds{1}$        & Value of indicator function. $\mathds{1}_{b,m}$ represents the value of indicator function corresponding to $b$-th example and $m$-th head. \\\midrule
        $\mathcal{L}$       & Training loss. $\mathcal{L}_l$ and $\mathcal{L}_u$ are the training loss for label examples and unlabeled examples respectively. $\ell_{b,m}$ is the loss corresponding to $b$-th example and $m$-th head.\\\midrule
        $\mathop{\arg\max}$ & Choose the most confident predicted class. \\\bottomrule
    \end{tabularx}
\end{table}

\section{Experimental setup}
We implement our experiments in PyTorch 1.6\footnote{\url{https://github.com/pytorch/pytorch}}. 
\subsection{Details of transformations}
The weak image transformation function applies random horizontal flip and random crop.
The one used for SVHN is slightly different from other datasets, as shown in Table~\ref{weakaug}.
For CIFAR-10, CIFAR-100, the operations include random horizontal flip followed by random crop.
Operations, such as horizontal flip, could be wrong variants for asymmetrical digit images, so only random crop is used for SVHN.
Note that one may further remove crop and translate transformation because these may transform the number 8 to number 3.
For convenience, we don't do it.

\begin{table}[H]
    \centering
    \begin{tabular}{llll}
        \toprule
        Operation        & Range            \\ \midrule
        Horizontal Flip  & [0,1]            \\
        Crop             & [-0.125,0.125]   \\ \bottomrule
    \end{tabular}
    \caption{
        List of operations for weak transformations of Multi-Head~Co-Training on different datasets. For SVHN, horizontal flip is skipped.
    }
    \label{weakaug}
\end{table}

The strong transformation is a modified version of RandAugment \cite{cubuk2020randaugment} followed by Cutout \cite{devries2017improved}.
The operations of RandAugment are shown in Table~\ref{strongaug}.
The range is similar to the original version, so we don't elaborate their meaning here.
Cutout randomly masks a square (with a side of length range from 0 to 0.5×image length) of pixels to gray. 

\begin{table}[H]
    \centering
    \begin{tabular}{ll}
        \toprule
        Operation      & Range        \\ \midrule
        AutoContrast   & [0, 1]       \\
        Brightness     & [0.05, 0.95] \\
        Color          & [0.05, 0.95] \\
        Contrast       & [0.05, 0.95] \\
        Equalize       & [0, 1]       \\
        Identity       & [0, 1]       \\
        Posterize      & [4, 8]       \\
        Rotate         & [-30, 30]    \\
        Sharpness      & [0.05, 0.95] \\
        ShearX         & [-0.3, 0.3]  \\
        ShearY         & [-0.3, 0.3]  \\
        Solarize       & [0, 256]     \\
        TranslateX     & [-0.3, 0.3]  \\
        TranslateY     & [-0.3, 0.3]  \\ \bottomrule
    \end{tabular}
    \caption{
        List of operations for strong transformations of the modified RandAugment. 
        Three transformations are randomly chosen and performed with stochastic magnitude.
    }
    \label{strongaug}
\end{table}

\subsection{Hyper-parameters}
We report the hyper-parameters for Multi-Head~Co-Training with three heads. 
Basically, the setting follows FixMatch \cite{sohn2020fixmatch} as shown in Table~\ref{hyperparameters}.
All of them stay the same across datasets unless otherwise stated.
For parameters updating, we use SGD with Nesterov momentum, weight decay, and cosine learning rate decay \cite{loshchilov2016sgdr}. 
The learning rate decay follows:
\begin{equation}
    \text{learning rate}\cdot \cos (\frac{7\pi \cdot \text{current iteration}}{16 \cdot \text{total iteration}})
\end{equation}
For CIFAR-10 and SVHN, WRN~28-2 is used as the backbone.
We trained the network on one single NVIDIA V100 for about 90 hours.
For CIFAR-100, the widen factor is adjusted to 8.
In Multi-Head~Co-Training, the number of channels of the final block in WRN~28-8 is changed from 512 to 256.
We trained the network on one single NVIDIA V100 for about 200 hours.
For Mini-ImageNet, we trained the network on one single NVIDIA V100 for about 30 hours.


\begin{table*}
    \centering
    \begin{tabular}{lccc}
        \toprule
        Hyper-parameters               & CIFAR-10, SVHN          &  CIFAR-100               &  Mini-ImageNet          \\ \midrule
        Batch size of labeled data     & 64                      &   64                     &   64                    \\ 
        Batch size of unlabeled data   & 448                     &   448                    &   192                   \\ 
        Iterations                     & $\text{2}^{\text{20}}$  &   $\text{2}^{\text{19}}$ &   300000                \\ 
        Weight of unsupervised loss    & 1                       &   1                      &   1                     \\
        Learning rate                  & 0.3                     &   0.3                    &   0.4                   \\ 
        momentum for learning rate     & 0.9                     &   0.9                    &   0.95                  \\ 
        Weight decay                   & 0.0005                  &   0.001                  &   0.0002                \\ \bottomrule
    \end{tabular}
    \caption{Multi-Head~Co-Training's hyper-parameters.}
    \label{hyperparameters}
\end{table*}

\section{Calibration}
Reliability diagram, confidence diagram, and Expected Calibration Error (ECE) are used in Section~\nameref{Calibration of SSL}. 
Predictions on the test set are divided into bins based on the confidence score (i.e., the maximum softmax probability).
The reliability diagram shows the average accuracy of the examples in each bin. 
So the gap between the actual average accuracy and the expected accuracy reflects whether the model is calibrated properly.
The confidence diagram shows how many examples are in each bin. 
ECE is the average over the absolute difference between accuracy and confidence,
\begin{equation}
    ECE=\sum^B_{b=1}\frac{|S_b|}{S}|acc(S_b)-conf(S_b)|
\end{equation}
where $S$ is the entire dataset and $S_b$ refers to the examples in $b$-th bin, $B$ is the number of bins. 
For convenience, $acc(b)$ refers to the average accuracy for examples in $b$-th bin and $conf(b)$ refers to the average confidence for examples in $b$-th bin.
As a one of the calibration methods, temperature scaling \cite{guo2017calibration} introduce a scale parameter $T$ in the softmax calculation, 
\begin{equation}
    q=\max_c  \frac{\exp(z_c/T)}{\sum_{j=1}^C\exp(z_j/T)}
\end{equation}
where $C$ represents number of class. We set $T=2$. 

\end{appendices}

\end{document}